\documentclass{article}
\usepackage{spconf,amsmath,graphicx}
\usepackage{hyperref}
\usepackage{cite}
\usepackage{subfig}
\usepackage{lipsum} 
\usepackage{changepage}
\usepackage[skip=5pt]{caption} %less space before caption text

\usepackage{stfloats}
\usepackage{xspace}
\xspaceaddexceptions{]\}}
\makeatletter
\DeclareRobustCommand\onedot{\futurelet\@let@token\@onedot}
\def\@onedot{\ifx\@let@token.\else.\null\fi\xspace}
 
\def\ie{\emph{i.e}\onedot}

\def\etal{\emph{et al}\onedot}
\makeatother

\title{End-to-end Multiple Instance Learning with Gradient Accumulation}
\name{Axel Andersson$\dagger$, Nadezhda Koriakina$\dagger$,
Nata\v{s}a Sladoje, Joakim Lindblad\thanks{\noindent$\dagger$ These authors contributed equally to this work; This work is supported by: Sweden’s Innovation Agency (VINNOVA), grants 2017-02447 and 2020-03611, the Swedish Research Council, grant 2017-04385, and European  Research Council, grant ERC‐2015‐CoG  682810}}
\address{Centre for Image Analysis, Dept. of Information Technology, Uppsala University, Sweden}

\usepackage{fancyhdr}
\fancypagestyle{FirstPage}{
\cfoot{\footnotesize{© 2022 IEEE.  Personal use of this material is permitted. Permission from IEEE must be obtained for all other uses, in any current or future media, including reprinting/republishing this material for advertising or promotional purposes, creating new collective works, for resale or redistribution to servers or lists, or reuse of any copyrighted component of this work in other works.}}
}

\begin{document}
\thispagestyle{FirstPage}
\maketitle

\begin{abstract}
Being able to learn on weakly labeled data, and provide interpretability, are two of the main reasons why attention-based deep multiple instance learning (ABMIL) methods have become particularly popular for classification of histopathological images. Such image data usually come in the form of gigapixel-sized whole-slide-images (WSI) that are cropped into smaller patches (instances). However, the sheer size of the data makes training of ABMIL models challenging.  All the instances from one WSI cannot be processed at once by conventional GPUs.  Existing solutions compromise training by relying on pre-trained models, strategic sampling or selection of instances, or self-supervised learning. We propose a training strategy based on gradient accumulation that enables direct \textit{end-to-end} training of ABMIL models without being limited by GPU memory.  We conduct experiments on both QMNIST and Imagenette to investigate the performance and training time, and compare with the conventional memory-expensive baseline and a recent sampled-based approach. This memory-efficient approach, although slower, reaches performance indistinguishable from the memory-expensive baseline.
\end{abstract}
\section{Introduction}
\label{sec:intro}
Deep learning based algorithms have been used to facilitate advancements in the field of medical image analysis. For example, in digital pathology, there has been an increased interest in computer-aided methods that can assist pathologists in making diagnosis~\cite{cheplygina2019not}. The data typically comes in the format of whole-slide-images (WSIs) and a common task is: Given a WSI, conclude what diagnosis a patient can have and what regions in the WSI are related to the diagnosis. Here, methods that do not rely on per-pixel annotations, but instead make use of a confirmed diagnostic information are preferable for several reasons: (i) per-pixel annotation is an expensive and laborious task; (ii) there is large inter- and intra-observer variability; (iii) fine annotations are more prone to human error, and (iv) usage of given fine ground truth does not facilitate knowledge discovery. Therefore, methods applied on such data with \textit{weak} (per-slide) labels are becoming increasingly popular, especially methods that fall under the category of multiple instance learning (MIL)~\cite{maron1998framework}. A common practice is to divide a WSI into patches, which are considered \textit{instances}, that compose a \textit{bag}, and to train a classifier that, given a bag of patches, predicts the diagnostic labels. Lately, MIL methods have been combined with attention based deep learning, forming the branch Attention Based Deep Multiple Instance Learning (ABMIL)~\cite{ilse2018attention}. These methods have become particularly popular as they can leverage the power of deep-learning, while simultaneously provide interpretability in the form of per-instance attention scores that indicate which instances are important for the classification of the WSI bag. Due to large number of patches per WSI, there is a practical challenge to apply deep-learning based algorithms on such data -- all the instances from one WSI/bag cannot be processed at once by GPU(s).  Existing solutions either separate the optimization of the feature encoder and the classifier~\cite{lu2021data,chen2021pan,li2021dual,dehaene2020self}, or rely on sub-sampling or strategic selection of instances to overcome these memory limitations~\cite{campanella2019clinical,chikontwe2020multiple,xie2020beyond,sharma2021cluster,koriakina2021effect}. 

We propose a training strategy, which does not require sub-sampling, for joint optimization of the classifier and the feature encoder. Our method relies on a type of gradient check-pointing strategy~\cite{chen2016training} that can be applied to ABMIL architectures. The method stems from the observation that the gradient of the  loss function with respect to the feature encoders parameters is separable and can therefore be computed in an accumulation-like fashion with very low requirements on GPU memory, thus enabling end-to-end training of ABMIL models also for very large data. This desired property comes at the price of a longer training time. The code is made open-source and publicly available at {\tt \href{https://github.com/axanderssonuu/ABMIL_with_accumulating_gradients}{github.com/axanderssonuu/ABMIL-ACC}}.

\section{Related work}
To circumvent the GPU memory limitations, methods where instances are first encoded into lower-dimensional feature representations, that easier fit on a GPU, have been proposed. For instance,~\cite{lu2021data, chen2021pan} rely on pre-trained networks for encoding the instances. Here, the encoders are fixed during training of the bag classifier. Consequentially, the performance of the classifier is limited by the quality of the feature encoders. Such limitations were addressed in~\cite{li2021dual,dehaene2020self}, where self-supervised techniques are proposed as an alternative. Although self-supervised methods may yield improved feature representations, they are still limited by strictly separating the optimization of the feature encoder and the classifier. Methods such as~\cite{campanella2019clinical, chikontwe2020multiple} managed to jointly train the feature encoder and classifier by strategically selecting few instances in the bag. Moreover,~\cite{xie2020beyond, sharma2021cluster, koriakina2021effect} enable joint optimization of the feature encoder and classifier by using sampling and/or clustering techniques. Although sampling enable training of both the feature encoder and classifier --- and can even lead to improved performance with some parameters --- it failed when the datasets contained too few sampled key instances per positive bag~\cite{koriakina2021effect}. In practice, it can be hard to know in advance the percent of key instances in real data and, consequently, to estimate the sample size. Pinckaers~\etal~\cite{pinckaers2021detection} also pointed out memory limitations related to computing the attention maps in ABMIL. Instead of using ABMIL,~\cite{pinckaers2020streaming,pinckaers2021detection} propose a regular convolutional neural network (CNN) for classification of the WSI. Here, the GPU memory limitations are avoided by dividing the whole-slide-image into smaller feasible tiles that sequentially can be processed by the CNN using a technique called streaming~\cite{pinckaers2020streaming} and gradient-check-pointing~\cite{chen2016training}.

\section{Method}
We start by introducing a few general concepts of ABMIL. We follow~\cite{ilse2018attention} and let $    \mathcal{B} = \{(x_1,y_1),\ldots,(x_n, y_n)\}$ denote a bag of instances $x_i \in \chi$, and their corresponding instance labels $y_i \in \{0,1\}$. The instance-level labels are not known in practice, only the label of the bag. Typically, the bag label is positive if a single instance label is positive. We let
\begin{equation*}
    y = 
    \begin{cases}
    0 \quad \text { if \quad} \sum_i^n y_i = 0 \\
    1 \quad \text{ otherwise}
    \end{cases}
\end{equation*}
denote the label of the bag. The objective of ABMIL is to train a model that, given a bag of instances, can predict the bag's label. The model consists of: (i) a feature encoder $z_i = f_\theta (x_i)$,
parameterized by the weights $\theta$, that transforms an individual instance $x_i$ into some feature representation $z_i$, and (ii) an attention-pooling classifier $y' = g_\varphi(\mathcal{Z})$,
parameterized by the weights $\varphi$, that pools all feature representations $\mathcal{Z} = \{z_1,\ldots,z_n\}$ and predicts a score, $y' \in [0,1]$ for the bag-label. The loss of scoring a bag $y'$ given the true label $y$ is computed using a loss function $\mathcal{L}(y,y')$. The conventional way of training ABMIL models is through variants of stochastic gradient descent (SGD). The crucial step in SGD is to compute the gradient of the  objective-function, $\mathcal{L}$, with respect to the model's parameters, $\theta$ and $\varphi$. To avoid redundant calculations, the gradient is computed in two phases: (i) A feed-forward phase where a bag of instances is fed through the network that predicts a bag-label. Feature activations for each of the instances are stored in memory to be used in the next phase. (ii) A back-propagation step where the gradient of the objective function is computed with respect to all weights in the network using the chain rule. The process is done iteratively, utilizing the stored activations from (i) to avoid redundant calculations. The drawback with this strategy is that activations from all instances in the bag are stored in computer memory simultaneously, quickly filling it up.

Our training strategy builds on the observation that gradient of the loss with respect to the parameters of the feature extractor can be decomposed into several terms. Specifically,
\begin{equation}
    \frac{\partial \mathcal{L}}{\partial \theta} = \sum_{i=1}^n \frac{\partial \mathcal{L}}{\partial g_\varphi (\{z_i\} \cup \mathcal{Z}_{\backslash i})} \frac{\partial g_\varphi(\{z_i\} \cup \mathcal{Z}_{\backslash i})}{\partial z_i} \frac{\partial z_i}{\partial \theta},
    \label{eq:1}
\end{equation}
where $\mathcal{Z}_{\backslash i} = \{z_1,\ldots,z_n\} \backslash \{z_i\}.$ This suggests that $\frac{\partial \mathcal{L}}{\partial \theta}$ can be computed with low requirements on GPU by using a type of gradient-check-pointing technique. If we treat $\mathcal{Z}_{\backslash i}$ as a constant, obtained by pre-evaluating $f$ on corresponding instances, the gradient in Eq.~\eqref{eq:1} can be computed without simultaneously storing in memory activations from all instances. In the extreme case we can simply (i) take a single instance $x_i$, (ii) feed it through $f$ while in training mode, \ie, so that activations are stored and available for back-propagation, (iii) concatenate the output with $\mathcal{Z}_{\backslash i}$, (iv) feed the concatenations $\{z_i\} \cup \mathcal{Z}_{\backslash i}$ through $g_\varphi$ (also in training mode), (v) compute the loss and (vi)  back-propagate. This effectively computes one term in Eq.~\eqref{eq:1}. Repeating it for all instances and accumulating the terms would eventually yield the true gradient. Importantly, with this gradient-accumulation strategy, the gradient computations are not limited by the total size of a bag, but merely on the size of a single instance. In a practice, it more efficient to use small batches of instances for training.  

The gradient of the loss with respect to the attention-pooling classifier's parameters, $\varphi$, is straightforward to compute. Since the feature representations of the instances are of a much lower dimension than the instances themselves, they should to some limit fit the memory of a GPU, and we can compute $\frac{\partial \mathcal{L}}{\partial \varphi}$ by directly forwarding $\mathcal{Z}$ through $g_\varphi$ and back-propagate.

\section{Experiments}
In a purely deterministic setting, the gradient accumulation strategy should result in a gradient equivalent to as if we could train with the entire bag at once. However, inconsistencies may appear due to numerical errors. Furthermore, many deep learning architectures include batch-normalizing layers. The effect of these layers depend on the number of instances that are simultaneously being forwarded through a network. Since the proposed training strategy breaks up a bag of instances into smaller \textit{batches} of instances, batch-normalization may lead to inconsistencies.

We construct a series of experiments, using synthetic bags from the publicly available datasets QMNIST~\cite{NEURIPS2019_51c68dc0} and Imagenette~\cite{imagewang}, to further investigate if training with the proposed gradient accumulation strategy is equivalent to training with the full bag.
\begin{figure}[t]
    \centering
    \includegraphics[width=0.48\textwidth]{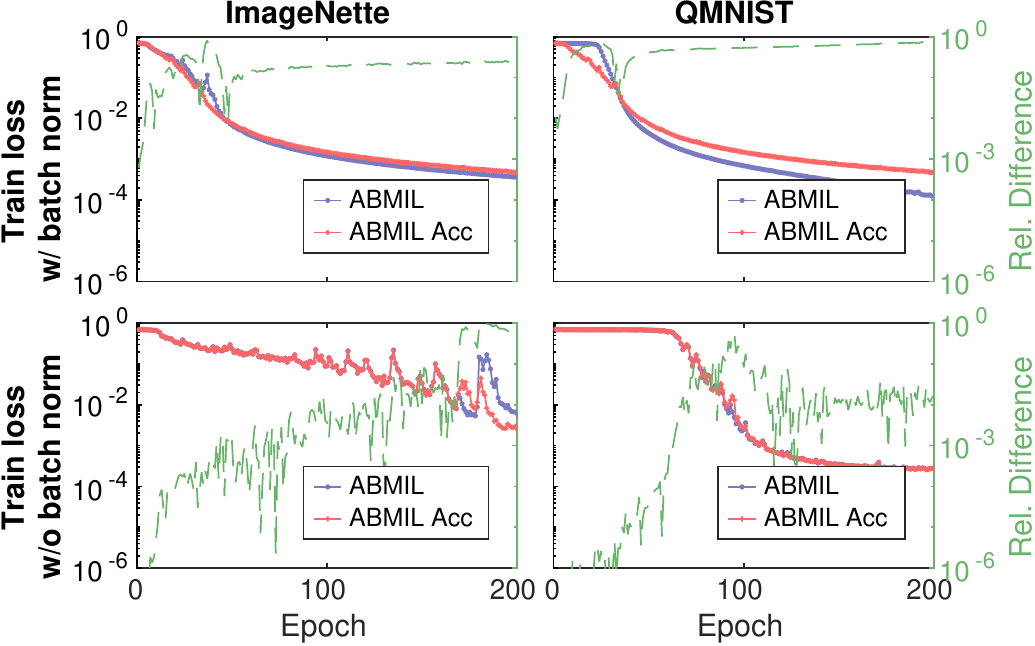}
    \caption{Comparison of the training loss of the proposed training strategy, ABMIL-ACC-25\%, with the training loss of conventional training (ABMIL), to see if the two strategies converge to the same solution. With (first row) and without (second row) batch normalizing layers.}
    \label{fig:convergence}
\end{figure}

% \subsection{QMNIST-bags}
\textbf{QMNIST-bags}
The original QMNIST dataset consists of train and test sets, each with 60k images. 
We randomly withdraw one third of the original QMNIST test set as a validation set, keeping the other two thirds as a test set. The train set is left unchanged. QMNIST-bags train, validation and test sets are sampled from these QMNIST train, validation and test sets correspondingly. We create each QMNIST-bags dataset with bags containing an equal number of instances per bag and where each instance appears at most once within a bag. Key instances are images of the digit ``9''. We perform experiments with datasets containing 100, 30 and 60 bags for training, validation and test respectively, 500 instances per bag and 5\% of key instances per positive bag. The images are resized to $84\!\times\!84$ pixels using bilinear interpolation.

% \subsection{Imagenette-bags}
\textbf{Imagenette-bags}
 We generate Imagenette-bags dataset from the Imagenette dataset containing train and test sets of $9479$ and $3935$ images correspondingly. We withdraw images from the original train set to compose a validation set that constitutes 20\% of all the training data and a new train set from remaining train images. Imagenette-bags train, validation and test sets are sampled from such train, validation and test sets correspondingly. During the creation of Imagenette-bags we resize the images to the size of $112\!\times\!112$ pixels using bilinear interpolation and perform a set of random augmentations (see \cite{koriakina2021effect} for details) using Albumentations package \cite{info11020125}.
Images of the class ``golf ball'' are selected as key instances. We conduct experiments with the created dataset containing 300 bags for training, 60 bags for validation and test, 150 instances per bag and 30\% of key instances per positive bag.

\subsection{Implementation Details}
For all of our experiments, we choose as the feature encoder, $f_\theta$, a small residual network (ResNet18), and the attention-pooling classifier, $g_\varphi$, similar to the non-gated attention pooling classifier in~\cite{li2019attention}. Learning rate is set to $5\cdot10^{-5}$, optimizer is used as in~\cite{koriakina2021effect} for QMNIST-bags and weight decay is set to $10^{-3}$. We observe the moving average error with a window of 15 epochs on the validation set, find the window that has the lowest average error, and save the model with the lowest validation error within this window.
The total number of training epochs is chosen observing convergence of train and validation errors in preliminary experiments; we let datasets to train for 300 epochs. 
We denote models trained using the proposed accumulation of gradients as ABMIL-ACC-$\alpha\%$, where $\alpha$ is the percent of instances in a bag that we feed together through our model.

\section{Results}
\subsection{Convergence comparison}
\label{sec:convergence}
First we compare convergence of a model trained using our gradient accumulation strategy  (ABMIL-ACC-25\%)  versus a model which uses conventional ABMIL training (\ie, ABMIL-ACC-100\%, the whole bag on GPU memory).  Fig~\ref{fig:convergence} presents the training loss, binary cross entropy, of a model trained using ABMIL-ACC and regular ABMIL, as well as the relative difference in loss between the models. Here, both models are initialized with the same weights and contain batch-normalizing layers. As shown, the convergence of the two methods is not identical. This discrepancy is due to batch normalizing layers.  In ABMIL-ACC, we accumulate the gradient by feeding small batches of instances through the network, resulting in slightly different normalization and a different convergence than regular ABMIL. We repeat the same experiment without batch normalizing layers, see Fig~\ref{fig:convergence} (bottom row). Here, the loss decline is essentially identical during the first $10$ epochs; the relative difference is less than $10^{-6}$, which is close to machine epsilon for single floating-point precision. As the number of epochs increases, numerical errors accumulate to such an extent that the two methods eventually start to converge to different solutions. It is therefore of interest to also evaluate the \textit{quality} of the solutions. 

\begin{figure}
    \centering
    \includegraphics[width=0.41\textwidth]{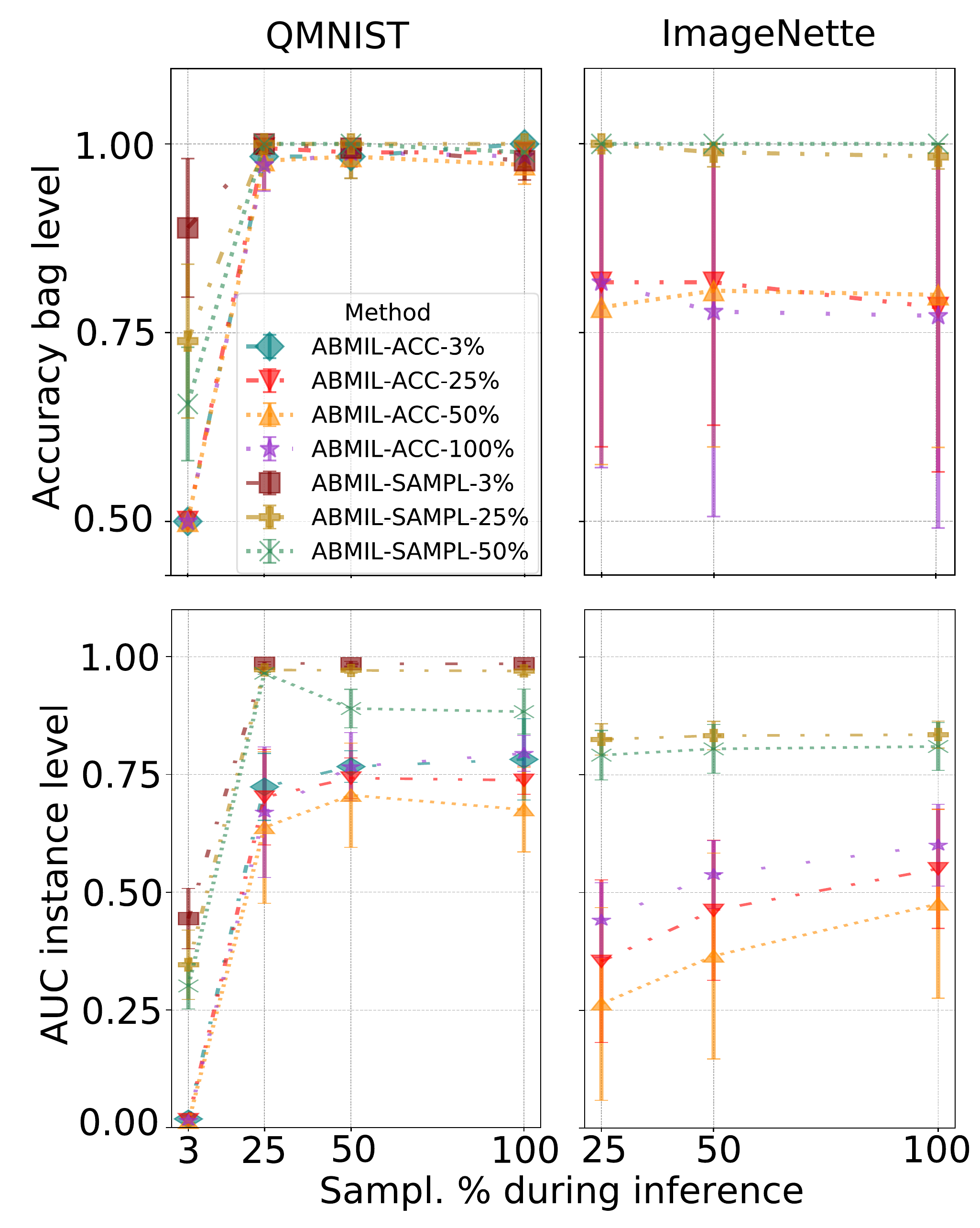}
    \caption{
    Test set performance on QMNIST-bags and Imagenette-bags (larger is better). \textbf{Top:} Accuracy at the bag level.
    \textbf{Bottom:} AUC at the instance level. 
    % Test set performance on QMNIST-bags with 100 bags for training, 500 instances per bag and 5\% of key instances and Imagenette-bags with 300 bags for training, 150 instances per bag and 30\% of key instances per positive bag (larger is better). 
    % \textbf{Top:} Accuracy at the bag level.
    % \textbf{Bottom:} AUC at the instance level.
    \vspace*{-2mm}}
    \label{fig:metrics_ALL}
\end{figure}

\subsection{Quantitative comparison}
Observing discrepancies in convergence due to batch normalization computations and rounding errors, we want to further investigate if the solutions obtained by accumulating the gradients are qualitatively different from those using the conventional ABMIL. To investigate this we consider the Imagenette and QMNIST datasets again. 
We repeat experiments 3 times, resampling datasets each time and training models from scratch.
To also evaluate the wall-clock training time, we train models with different constraints on GPU memory by changing the $\alpha$ parameter. We perform experiments with $\alpha\%$ equal to 25, 50 and 100 (where $\alpha=100$ corresponds to conventional ABMIL) for both datasets and additional $\alpha\%$ equal to 3 for QMNIST.

\begin{figure}
    \centering
    \subfloat[QMNIST-bags]{\includegraphics[width=0.44\textwidth,trim={1mm 2mm 1mm 1mm},clip]{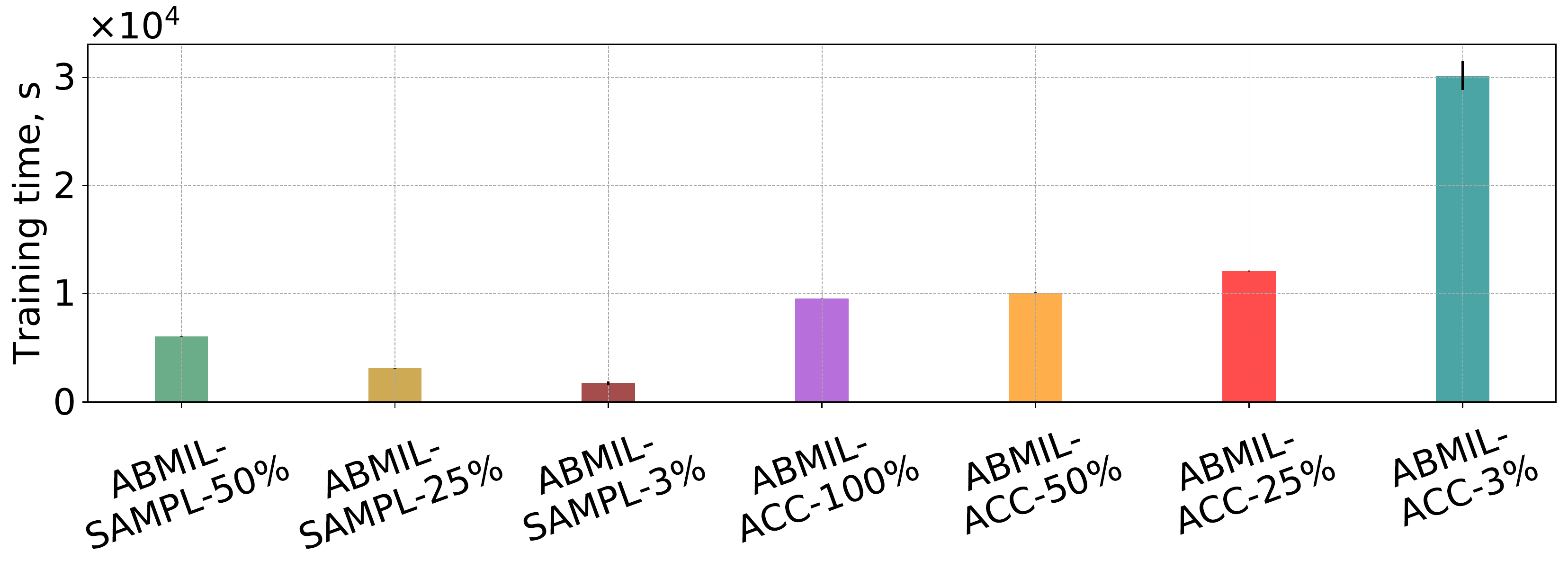}\label{fig:time_qmnist}}\\ \vspace{-10pt}
    \subfloat[Imagenette-bags]{\includegraphics[width=0.44\textwidth,trim={1mm 2mm 1mm 1mm},clip]{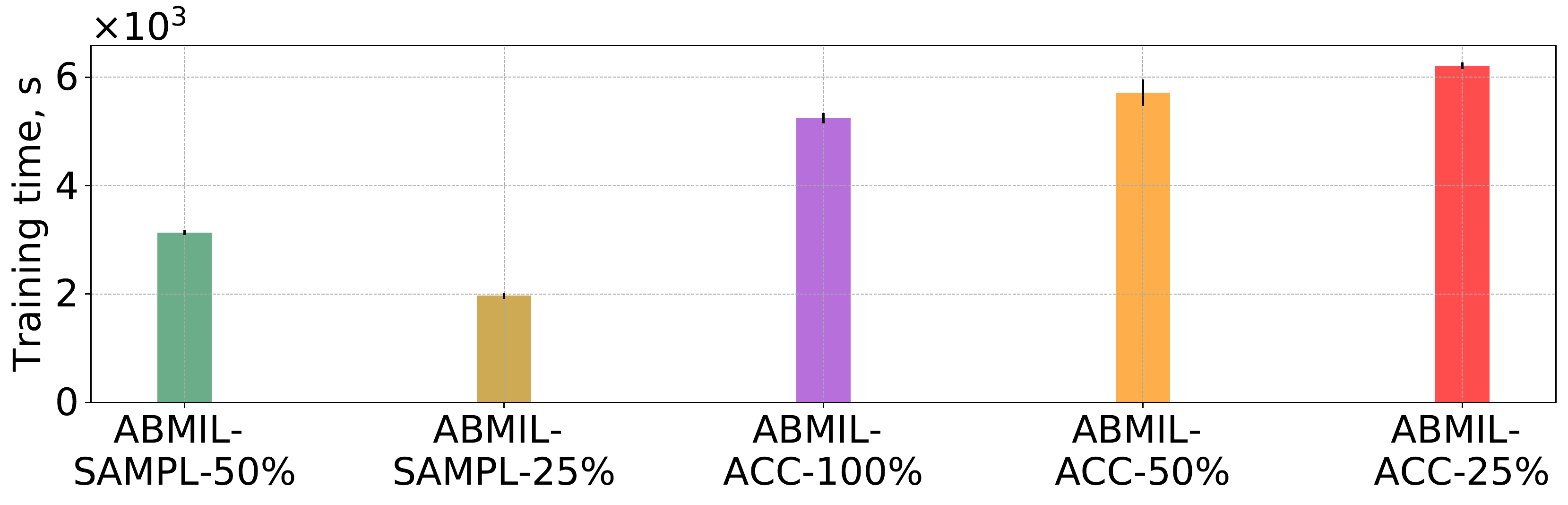}\label{fig:time_Imagenette}}
    \caption{Time in seconds for training QMNIST-bags (top) and Imagenette-bags (bottom) models using different strategies.\label{fig:time}}
\end{figure}

Lastly, we compare bag-level accuracy, instance-level AUC~\cite{koriakina2021effect}, and training time, between models trained using ABMIL-ACC, baseline ABMIL models trained using conventional training, as well as models trained using the sampling strategy presented in~\cite{koriakina2021effect}. The method in~\cite{koriakina2021effect} performs sampling both during training and during inference. We chose to decouple the sampling during training and sampling during inference to study their effect separately and apply inference sampling on ABMIL-ACC variants too. The results are presented in Fig~\ref{fig:metrics_ALL} for QMNIST-bags and Imagenette-bags. All the variants reach high accuracy at the bag level unless inference sampling percent is 3 (for QMNIST-bags). Unsurprisingly, the variants of ABMIL-ACC with harder constraints on GPU memory (ABMIL-ACC-3\% and ABMIL-ACC-25\%) require longer time to train (see Fig~\ref{fig:time}), but converges to a solution that is of the same quality (bag level accuracy, AUC instance level) as the memory expensive original ABMIL method. 

Previously developed method ABMIL-SAMPL outperforms other methods in terms of AUC at the instance level when sampling percent during training is 3, 25 or 50 and inference sampling percent is 25 and higher. The effect of sampling during inference is not prominent for current datasets except for the case of 3\% (see Fig~\ref{fig:metrics_ALL}), where performance drops. Thus, we can conclude that sampling with low percentage is beneficial for training but detrimental for testing. 

\section{Discussion and conclusion}
By accumulatively computing the gradients we avoid the otherwise often prohibitively expensive GPU-memory requirements of ABMIL and can train models end-to-end. The proposed strategy proved slower in terms of wall-clock training time, while it converges to a solution similar to the conventional, GPU-expensive, way of training. As discussed in Section~\ref{sec:convergence}, the discrepancy in solutions can be explained by differences in batch-normalizing layers and accumulation of numerical errors.  

Surprisingly, the instance level AUC for models trained using regular ABMIL or accumulation of gradients, ABMIL-ACC, are significantly lower than models trained using the sampling strategy in~\cite{koriakina2021effect}. Intuitively, one may expect that training without sub-sampling should lead to a better performing model. We speculate that conventional training might suffer from a type of ``feedback-loop'' phenomenon. Early during training, the model decides to pick up on some characteristic feature. The model's parameters are then updated so that similar types of features obtain more attention in consecutive epochs. Possibly, sub-sampling instances from a bag introduces variations (a bit like drop-out) leading to a more robust solution. This phenomenon, along with alternative approaches for attention-pooling, is something we want to investigate in the future.

\vfill\pagebreak
% \newpage

\bibliographystyle{IEEEbib}
\bibliography{strings,refs}

\end{document}